\documentclass[11pt,a4paper]{article}
\usepackage[hyperref]{emnlp2020}
\usepackage{times}
\usepackage{latexsym}

\usepackage{microtype}

\aclfinalcopy 

\usepackage[normalem]{ulem}
\usepackage{multirow}
\usepackage{makecell}
\usepackage{soul}
\usepackage{enumitem}
\usepackage{adjustbox}
\usepackage{amsmath}
\usepackage{amssymb}
\usepackage{array}
\usepackage{graphicx}
\usepackage{comment}
\usepackage{booktabs}
\usepackage{xcolor,colortbl} 
\usepackage{algorithm}
\usepackage{algorithmic}
\newcommand{\tabincell}[2]{\begin{tabular}{@{}#1@{}}#2\end{tabular}}

\title{Dialogue Generation on Infrequent Sentence Functions \\ via Structured Meta-Learning}

\author{Yifan Gao\textsuperscript{1}, Piji Li\textsuperscript{2}, Wei Bi\textsuperscript{2}, Xiaojiang Liu\textsuperscript{2}, Michael R. Lyu\textsuperscript{1}, and Irwin King\textsuperscript{1}   \\
	{\textsuperscript{1} Department of Computer Science and Engineering, }\\
	{The Chinese University of Hong Kong, Shatin, N.T., Hong Kong}\\
	{\textsuperscript{2} Tencent AI Lab, Shenzhen, China}\\
    { \textsuperscript{1}\{yfgao,king,lyu\}@cse.cuhk.edu.hk}
	{ \textsuperscript{2}\{pijili,victoriabi,kieranliu\}@tencent.com}
}

\date{}

\begin{document}
\maketitle
\begin{abstract}
Sentence function is an important linguistic feature indicating the communicative purpose in uttering a sentence.
Incorporating sentence functions into conversations has shown improvements in the quality of generated responses.
However, the number of utterances for different types of fine-grained sentence functions is extremely imbalanced.
Besides a small number of high-resource sentence functions, a large portion of sentence functions is infrequent.
Consequently, dialogue generation conditioned on these infrequent sentence functions suffers from data deficiency.
In this paper, we investigate a structured meta-learning (SML) approach for dialogue generation on infrequent sentence functions.
We treat dialogue generation conditioned on different sentence functions as separate tasks, and apply model-agnostic meta-learning to high-resource sentence functions data. 
Furthermore, SML enhances meta-learning effectiveness by promoting knowledge customization among different sentence functions but simultaneously preserving knowledge generalization for similar sentence functions. 
Experimental results demonstrate that SML not only improves the informativeness and relevance of generated responses, but also can generate responses consistent with the target sentence functions.
\end{abstract}

\section{Introduction}
\begin{figure}[t!]
\centering
\includegraphics[width=1.0\columnwidth]{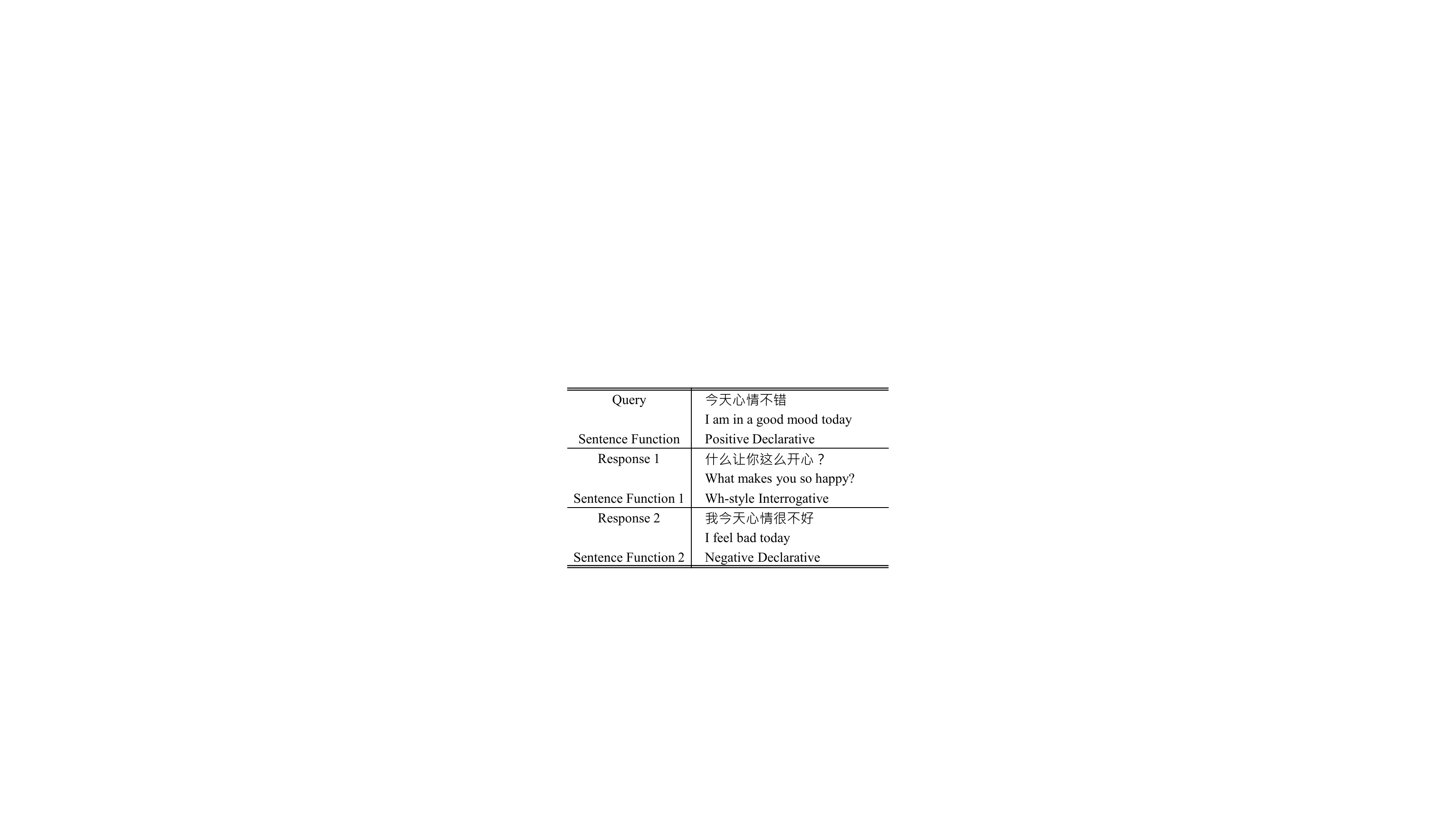}
\caption{Query-response pairs with fine-grained sentence functions. Responses under different sentence functions are completely different in global structures.}
\label{fig:example}
\end{figure}

Humans express intentions in conversations through sentence functions, such as interrogation for acquiring further information, declaration for making statements, and imperative for making requests and instructions.
For machines to interact with humans, it is therefore essential to enable them to make use of sentence functions for dialogue generation.
Sentence function is an important linguistic feature indicating the communicative purpose of a sentence in a conversation.
There are four major sentence functions: \textit{Declarative}, \textit{Interrogative}, \textit{Exclamatory} and \textit{Imperative} \cite{Rozakis2003CompleteIG}.
Each major sentence function can be further decomposed into fine-grained ones according to different purposes indicated in conversations.
For example, \textit{Interrogative} is divided into \textit{Wh-style Interrogative}, \textit{Yes-no Interrogative} and other types.
These fine-grained sentence functions have great influences on the structures of utterances in conversations including word orders, syntactic patterns, and other aspects \cite{akmajian1984sentence,yule2016study}.
Figure \ref{fig:example} presents how sentence functions influence the responses. 
Given the same query expressed in \textit{Positive Declarative}, the responses expressed in \textit{Wh-style Interrogative} and in \textit{Negative Declarative} are completely different.

Although the use of sentence functions improves the overall quality of generated responses \cite{Ke2018GeneratingIR}, it suffers from the data imbalance issue.
For example, in the recently released response generation dataset with manually annotated
sentence functions STC-SeFun \cite{Bi2019FineGrainedSF}, more than 40\% of utterances are \textit{Positive Declarative} while utterances annotated with \textit{Declarative with Interrogative words} account for less than 1\%  of the entire dataset.
Therefore, dialogue generation models suffer from data deficiency for these infrequent sentence functions.

Recently, model-agnostic meta-learning (MAML)~ \cite{Finn2017ModelAgnosticMF} has shown promising results on several low-resource natural language generation (NLG) tasks, including neural machine translation \cite{Gu2018MetaLearningFL}, personalized response generation \cite{Madotto2019PersonalizingDA} and domain-adaptive dialogue generation \cite{Qian2019DomainAD}.
They treat languages of translation, personas of dialog and dialog domains as separate tasks in MAML respectively.
In the same spirit of previous works, we first treat dialogue generation conditioned on different sentence functions as separate tasks, and meta-train a dialogue generation model using high-resource sentence functions.
Moreover, we observe that sentence functions have hierarchical structures: four major sentence functions can be further divided into twenty fine-grained types.
Some fine-grained sentence functions may share some similarities while some others are disparate.
For example, utterances belong to \textit{Wh-style Interrogative} and \textit{Yes-no Interrogative} may share some transferable word patterns while utterances in \textit{Wh-style Interrogative} and in \textit{Exclamatory with interjections} totally differ from each other. 
Motivated by this observation, we explore a structured meta-learning (SML) considering inherent structures among fine-grained sentence functions.
Inspired from recent advances on learning several initializations with a set of meta-learners \cite{Yao2019HierarchicallySM,Vuorio2019MultimodalMM}, we develop our own approach to utilize the underlying structure of sentence functions.
More specifically, our proposed SML explicitly tailors transferable knowledge among different sentence functions.
It utilizes the learned representations of fine-grained sentence functions as parameter gates to influence the globally shared parameter initialization.
Therefore, conversation models for similar sentence functions can share similar parameter initializations and vice versa.
As a result, SML enhances meta-learning effectiveness by promoting knowledge customization among different sentence functions but simultaneously preserving knowledge generalization for similar sentence functions.

The experimental results on STC-SeFun dataset \cite{Bi2019FineGrainedSF} show that responses generated from our proposed structured meta-learning algorithm are of better quality over several baselines in both human and automatic evaluations. 
Moreover, our proposed model can generate responses consistent with the target sentence functions while baseline models may ignore the target sentence functions or generate some generic responses.
We further conduct a detailed analysis on our proposed model and show that it indeed can learn word orders and syntactic patterns for different fine-grained sentence functions.

\section{Background}
\paragraph{Dialogue Generation with Sentence Function.}
Open domain dialogue generation has been widely studied with sequence-to-sequence learning (Seq2Seq) \cite{Sutskever2014SequenceTS}.
To alleviate the generate generic and dull responses issue of Seq2Seq \cite{li-etal-2016-diversity},
some efforts provide additional controlling signals in dialogue generation, such as emotion \cite{Zhou2017EmotionalCM}, persona \cite{Zhang2018PersonalizingDA}, and topic \cite{Xing2016TopicAN}.
Different from these local controlling factors, sentence function can influence the global structure of the entire response such as changing word orders and word patterns. 
\citet{Zhao2017LearningDD} utilize dialogue acts as prior linguistic knowledge and integrate it with conditional variational autoencoders to achieve the discourse-level diversity of generated responses.
\citet{Ke2018GeneratingIR} adopt a conditional variational autoencoder to capture various word patterns and introduce a type controller to control the sentence function. 
\citet{Xu2019NeuralRG} generalize the concept of dialogue act into meta words and use meta words for open domain dialogue generation.


\paragraph{Meta-Learning for Low-Resource NLG.}
Humans can learn quickly with a few examples while data-driven models are mostly compute-intensive. 
In meta-learning, the goal of the trained model is to quickly learn a new task from a small amount of data.
Therefore, the model should be able to learn transferable knowledge on a large number of different tasks.
Recently, model-agnostic meta-learning (MAML) \cite{Finn2017ModelAgnosticMF} has shown promising results on several few-shot classification tasks.
MAML directly optimizes the gradient towards a good parameter initialization for easy fine-tuning on low-resource scenarios.
Because of the model-agnostic nature of MAML, it can be directly applied to low-resource NLG tasks with modifications on corresponding training strategies. 
\citet{Gu2018MetaLearningFL} frame machine translation between two language pairs as a single task in meta-learning, and learn to adapt to low-resource languages based on multilingual high-resource language tasks.
In a similar spirit, recent works apply MAML to personalized response generation \cite{Madotto2019PersonalizingDA} and task-oriented dialogue agents \cite{Mi2019MetaLearningFL,Qian2019DomainAD}.
In this paper, we not only investigate how MAML helps for open domain dialogue generation on infrequent sentence functions, but also develop a structured approach to fit the hierarchical structure of sentence functions.

\section{Problem Formulation}
We define response generation conditioned on every query-response sentence function pair $(d_X, d_Y)$ as a single task. 
As the number of utterances for different sentence functions is extremely imbalanced, some tasks have abundant utterances while some others are low-resource.
We take $K$ high-resource tasks as training data, denoted as:
\begin{align}
\resizebox{0.89\hsize}{!}{$
D_{train}^k=\{(X^k_n, Y^k_n, d_X^k, d_Y^k), n = 1...N\}, k = 1...K $}
\end{align}
Then, we take $T$ tasks with infrequent sentence functions as target tasks, denoted as:
\begin{align}
\resizebox{0.89\hsize}{!}{$ D_{target}^t=\{(X^t_n, Y^t_n, d_X^t, d_Y^t), n = 1...N'\}, t = 1...T $}
\end{align}
where $N' \ll N$.

During training, taking the query $X$, its sentence function $d_X$ and the target response sentence function $d_Y$ as inputs, a dialog model $f$ parameterized by $\theta$ learns the mapping between inputs and the corresponding response $Y$ using training data $D_{train}$ of $K$ tasks,
\begin{align}\label{eqn:meta-train}
    f_{\theta}: X^k \times (d_X^k, d_Y^k) \rightarrow Y^k, k = 1...K
\end{align}

The initialization parameters of model $f$ learned from the training process, denoted by $\theta_0$, are used as the initialization parameters in the adaptation process. The adaptation process on each target task $D_{target}^t$ can be formulated as follows:
\begin{align}\label{eqn:meta-test}
    f_{\theta^*} = \arg\max_{\theta} \log p(f_{\theta} | D_{target}^t, f_{\theta_0})
\end{align}
where $f_{\theta^*}$ is the fine-tuned model that could perform well on the target task $D_{target}^t$.



\section{Proposed Approach}
In this section, we first introduce the conditional sequence-to-sequence (C-Seq2Seq) model for open domain dialogue generation with fine-grained sentence function.
Then we describe how to meta-train C-Seq2Seq under the algorithm of model-agnostic meta-learning.
Finally, we explore the structure of fine-grained sentence functions and propose the structured meta-learning (SML) algorithm.

\subsection{C-Seq2Seq} \label{sec:seq2seq}
Conditional sequence-to-sequence (C-Seq2Seq) \cite{Ficler2017ControllingLS} is the best generative model on STC-SeFun dataset \cite{Bi2019FineGrainedSF}.
We use it to test the effectiveness of our proposed structured meta-training approach.
C-Seq2Seq follows the widely used encoder-decoder framework \cite{Sutskever2014SequenceTS,Vinyals2015PointerN}.
The encoder transforms the query $X$ into contextualized representation $(\mathbf{h}_1, \mathbf{h}_2, ..., \mathbf{h}_n)$ through bidirectional LSTMs \cite{Hochreiter1997LongSM}.
For the decoder part, we learn an additional sentence function embedding $[\mathbf{s}_1, \mathbf{s}_2, ..., \mathbf{s}_K]$ for each query-response sentence function pair, which plays a major role in our structured modeling (Sec. \ref{sec:structure-maml}). 
Then we takes the concatenation of word embedding $\mathbf{w}_t$ and the sentence function embedding $\mathbf{s}_k$ as input at each timestep, and updates its hidden state as follows,
\begin{align}
    \mathbf u_t = \text{LSTM}(\mathbf u_{t-1}, [\mathbf w_{t}; \mathbf{s}_k]). 
\end{align}
The decoder utilizes soft attention mechanism \cite{Luong2015EffectiveAT} to derive the context vector $\mathbf{c}_t$,
\begin{align}
\resizebox{0.89\hsize}{!}{$
    a_{t,i} = \frac{\text{exp}(\mathbf{u}_t^\top \mathbf{W}_a \mathbf{h}_i)}{\sum_{j} \text{exp}(\mathbf{u}_t^\top \mathbf{W}_a \mathbf{h}_j)},  ~
    \mathbf{c}_t = \sum_i a_{t,i}\mathbf{h}_i. 
    $}
\end{align}
Finally, the predicted probability distribution over the vocabulary $V$ is computed as:
\begin{align}
\tilde{\mathbf{h}}_t  &= \text{tanh}(\mathbf{W}_{h} [\mathbf{u}_t;\mathbf{c}_t]), \\
\text{P}_{V} &= \text{softmax}(\mathbf{W}_V \tilde{\mathbf{h}}_{t} + \mathbf{b}_V), 
\end{align}
where $\mathbf{W}_a$, $\mathbf{W}_h$, $\mathbf{W}_V$ and $\mathbf{b}_V$ are trainable parameters.

\begin{figure*}[ht!]
\centering
\includegraphics[width=1.0\textwidth]{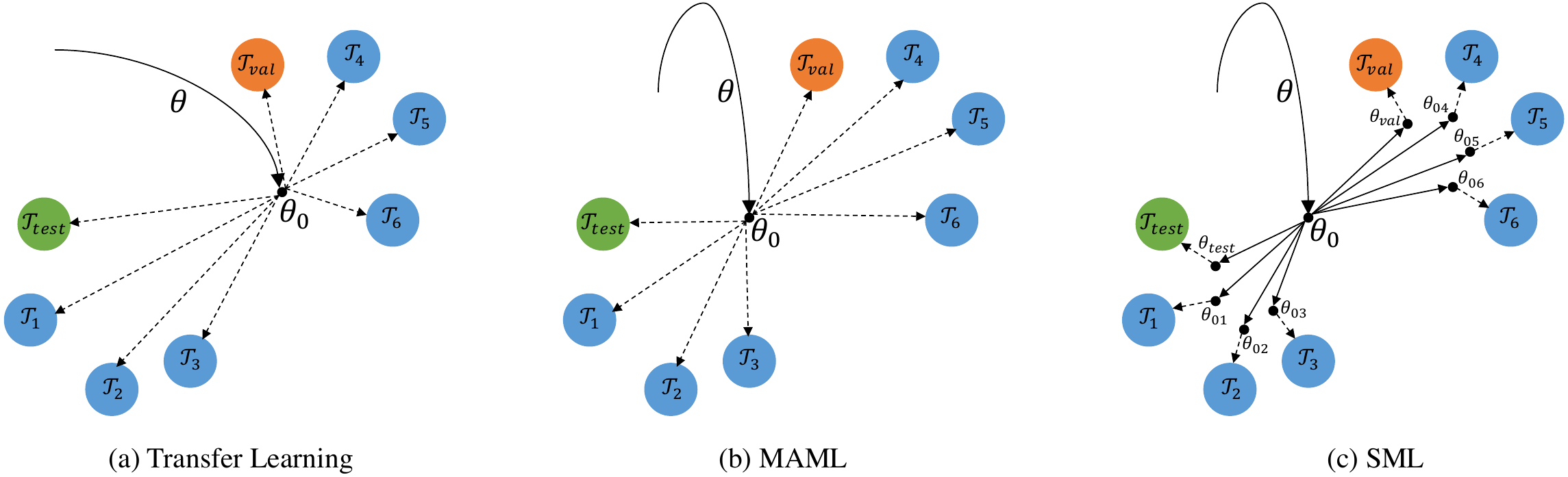}
\caption{An intuitive comparison among (a) Transfer Learning, (b) Model-Agnostic Meta-Learning (MAML) and (c) Structured Meta-Learning (SML). Tasks of training, validation and testing are colored in blue, orange and green respectively. Solid lines represent the learning of initialization $\theta_0$ while dashed lines show the path of fine-tuning.
Our structured modeling can learn the structure in different sentence functions so that similar tasks will be initialized from closer starting points than others ($T_{1,2,3}$ and $T_{4,5,6}$ in (c)). In the testing (adaptation) stage, a new sentence function such as Negative Declarative will benefit from this learned structure by initializing from a point that is close to other fine-grained sentence functions in the same category of Declarative ($T_{test}$ in (c)).
}
\label{fig:comparison}
\end{figure*}

\subsection{Meta-Learning for C-Seq2Seq} \label{sec:maml}

The fundamental idea behind meta-learning is based on a simple machine learning principle: test and train conditions must match.
In the context of meta-learning, it becomes that the conditions between task adaptation (fine-tuning) stage (Eqn.~\ref{eqn:meta-test}) and meta-training stage (Eqn.~\ref{eqn:meta-train}) must match.
To mimic the task adaptation stage, Model-Agnostic Meta-Learning (MAML) \cite{Finn2017ModelAgnosticMF} explicitly train the parameters of the model such that a small number of gradient steps with a small amount of training data will make rapid progress on new tasks.
The intuition behind MAML is that there exist some transferable internal representations across tasks. MAML aims to find the most sensitive model parameters such that small changes in model parameters will produce large improvements on each task.

Here is how we apply MAML to response generation on infrequent sentence functions. 
We uniformly sample one source task $\mathcal{T}_k$ at random. 
Then we independently sample two subsets of data $(D_{\mathcal{T}_k}, D_{\mathcal{T}_k}^{'})$ from task $\mathcal{T}_k$.
$D_{\mathcal{T}_k}$ is used to simulate the process when $f_\theta$ adapts to the target low-resource tasks while $D_{\mathcal{T}_k}^{'}$ is used to evaluate the outcome of the adapted model. 

In the simulation of adaptation stage, the model $f$ parameterized by $\theta$ adapts to this new task $\mathcal{T}_k$ using one or more gradient descent updates,
\begin{align}\label{eqn:inner-update}
    \theta_{k}^{'} = \theta - \alpha \nabla_{\theta} \mathcal{L}^{D_{\mathcal{T}_k}}(f_{\theta}),
\end{align}
where $\alpha$ is a hyperparameter for task-specific learning rate.
Then the model evaluates the updated parameters $\theta_{k}^{'}$ towards $D_{\mathcal{T}_k}^{'}$. The loss can be formulated as,
\begin{align}
    \mathcal{L}^{D_{\mathcal{T}_k}^{'}}(f_{{\theta}_{k}^{'}}) = \mathcal{L}^{D_{\mathcal{T}_k}^{'}}(f_{\theta - \alpha \nabla_{\theta} \mathcal{L}^{D_{\mathcal{T}_k}}(f_{\theta})})
\end{align}

Afterward, the model is trained by optimizing the performance of $\mathcal{L}(f_{{\theta}_{k}^{'}})$ with respect to $\theta$ across randomly sampled tasks. 
To learn the internal representation shared across tasks, it is possible to aggregate gradients $\nabla_{\theta}\mathcal{L}(f_{{\theta}_{k}^{'}})$ sampled from several tasks in the meta-update,
\begin{align} \label{eqn:meta-update}
    \theta \leftarrow \theta - \beta \sum_{k} \nabla_{\theta} \mathcal{L}(f_{{\theta}_{k}^{'}}),
\end{align}
where $\beta$ is the meta learning rate across tasks and $k$ is the sampled tasks for gradient aggregation\footnote{In our implementation, we follow \newcite{Finn2017ModelAgnosticMF} to adopt a first-order approximation for the meta-gradient update to reduce the computational complexity.}.
Different from common gradient-based approaches, Eqn.~\ref{eqn:meta-update} update the model not from $\theta_{k}^{'}$ but from $\theta$ because MAML aims to learn the most sensitive parameters to facilitate fast adaptation.
As a result, the meta-learned model is not necessarily a good model on its own, but it adapts fast on any new task with a few gradient update steps.

\subsection{Exploring Structure Modeling} \label{sec:structure-maml}
MAML learns some transferable knowledge in different training tasks (a task in our paper is defined as response generation conditioned on a given sentence function). In effect, the meta-learned model can adapt fast for the low-resource testing tasks (sentence functions). However, MAML assumes all tasks in training and adaptation stages distributed uniformly, which is not the case for our conditioned response generation -- some tasks may share some similarities while some are exclusive to each other.
For example, utterances belong to \textit{Wh-style Interrogative} and \textit{Yes-no Interrogative} may share some transferable word patterns while word patterns in \textit{Wh-style Interrogative} and \textit{Exclamatory with interjections} are totally different.
Therefore, we propose to represent sentence functions explicitly through learned embeddings $\mathbf{s}_1; ...; \mathbf{s}_K$ (Sec. \ref{sec:seq2seq}).
Then sentence function embeddings are used to interact with each other via a gated self-attention mechanism, which can be viewed as a clustering process to make similar sentence function embeddings close to each other.
Finally, the self-attended representations of these sentence functions are used as parameter gates to tailor the transferable knowledge of the meta-learned prior parameters.
Recently, the task-aware modulation problem in meta-learning is also investigated in the machine learning community \cite{Yao2019HierarchicallySM,Vuorio2019MultimodalMM}.
In their approaches, they first learn the task mode from input data as a vectorized representation and use the identified mode to modulate the meta-learned prior parameters.
The key difference between our structured modeling and their approaches is our model does not learn the task-aware representation through input data because the sentence functions are predefined before generating the responses.
Moreover, we propose the gated self-attentive approach to learn the underlying structure which has never been used before.

\paragraph{Task Representation Learning.}
To ensure similar tasks share similar representations, we design a gated self-attention module for sentence function embeddings.
For any fine-grained sentence function $k$,
we first match the sentence function embedding matrix $\mathbf{S} = [\mathbf{s}_1; ...; \mathbf{s}_K]$ with itself $\mathbf{s}_k$ to compute the self-matching representation $\mathbf{m}_k$, and then combine it with the original representation $\mathbf{s}_k$:
\begin{align} \label{eqn:selfattn1}
    \mathbf{a}_k &= \text{softmax(}{\mathbf{S}}^\top  \mathbf{s}_k \text{)}, 
    ~\mathbf{m}_k = \mathbf{S} \mathbf{a}_k \\
    \mathbf{f}_k &= \text{tanh(}\mathbf{W}_f[\mathbf{s}_k; \mathbf{m}_k]\text{)},
\end{align}
The self-matching operation matches the sentence form embedding to other sentence form embeddings, which can be viewed as a clustering process so that embeddings from similar sentence forms will be close to each other.

The final representation $\tilde{\mathbf{s}}_k$ is derived via a gated summation through a learnable gate vector $\mathbf{g}_k$,
\begin{align}
    \mathbf{g}_k &= \text{sigmoid(}\mathbf{W}_g[\mathbf{s}_k; \mathbf{m}_k]\text{)} \\
    \tilde{\mathbf{s}}_k &= \mathbf{g}_k \odot \mathbf{f}_k + (1-\mathbf{g}_k) \odot \mathbf{s}_k \label{eqn:selfattn2}
\end{align}
where $\mathbf{W}_{f}$, $\mathbf{W}_{g}$ are learnable weights, $\odot$ is the element-wise multiplication.
For sentence function embeddings $\mathbf{s}_{new}$ in the adaptation stage, we use the already well-learned sentence function embeddings $\mathbf{S} = [\mathbf{s}_1; ...; \mathbf{s}_K]$ in the meta-training stage, concatenate $\mathbf{s}_{new}$ with learned embeddings as $\mathbf{S}^{'} = [\mathbf{s}_1; ...; \mathbf{s}_K; \mathbf{s}_{new}]$ and apply Eqn.~\ref{eqn:selfattn1} $\sim$ \ref{eqn:selfattn2} for task representation learning in the adaptation stage. 
Because sentence form embeddings seen in training are learned to build the underlined structure: similar sentence functions are clustered close to each other. In the adaptation phase, the unseen sentence function embedding can adapt fast by moving to the cluster it belongs to. For example, a new sentence form Yes-no Interrogative can learn some transferable knowledge from trained sentence forms under the Interrogative category.

\paragraph{Task-Specific Knowledge Adaptation.}
To adapt globally transferable knowledge $\theta_0$ to each sentence function, we design a parameter gate $\mathbf{o}_k$ for $\theta_0$,
\begin{eqnarray} \label{eqn:adaptation} 
    \mathbf{o}_k = \text{FC}_{\mathbf{W}_p}^{\sigma}(\tilde{\mathbf{s}}_k), ~~~~~~
    \theta_{0k} = \theta_0 \odot \mathbf{o}_k 
\end{eqnarray}
where $\text{FC}_{\mathbf{W}_p}^{\sigma}$ is a fully connected layer parameterized by $\mathbf{W}_p$ and activated by a sigmoid function $\sigma$, $\odot$ is the element-wise multiplication.
Intuitively, sentence functions with similar representations will activate similar initial parameters while dissimilar sentence functions trigger different ones.
One major problem for Eqn.~\ref{eqn:adaptation} is that it introduces dozens of parameters compared to $\theta_0$ to achieve the element-wise dot product with $\theta_0$. 
Here we only tailor parameters in the decoder to reduce the total amount of learnable parameters.

\begin{algorithm}[t]
\fontsize{10}{13}\selectfont
\begin{algorithmic}[1]
	\REQUIRE
	{$\mathcal{E}$: distribution over tasks \{$\mathcal{T}_1, ..., \mathcal{T}_K$\}}
	\REQUIRE
	$\alpha, \beta$: step size hyperparameters
	\STATE Randomly initialize $\theta$
	\WHILE{not done} 
	\STATE Sample a batch of tasks $\mathcal{T}_k \sim \mathcal{E}$
	\FOR {all $\mathcal{T}_k$}
	\STATE Sample $D_{\mathcal{T}_k}, D_{\mathcal{T}_k}^{'}$ from $\mathcal{T}_k$
	\STATE Compute task representation $\tilde{\mathbf{s}}_k$ in Eqn.~\ref{eqn:selfattn2}
	\STATE Compute $\theta_{0k}$ in Eqn.~\ref{eqn:adaptation}
	\STATE Evaluate $\nabla_{\theta_{0k}} \mathcal{L}(f_{\theta_{0k}})$ with respect to $D_{\mathcal{T}_k}$
	\STATE Update $\theta_{0k}^{'} = \theta_{0k} - \alpha \nabla_{\theta_{0k}} \mathcal{L}(f_{\theta_{0k}})$ in Eqn.~\ref{eqn:inner-update}
	\ENDFOR
	\STATE Update 
	$\theta \leftarrow \theta - \beta \sum_{k} \nabla_{\theta_{0k}^{'}} \mathcal{L}(f_{{\theta}_{0k}^{'}})$ in Eqn.~\ref{eqn:meta-update} with respect to all $D_{\mathcal{T}_k}^{'}$
	\ENDWHILE
	\caption{Meta-training of SML}
	\label{alg:sml}
\end{algorithmic}	
\end{algorithm}

\paragraph{SML Algorithm and Visualization.}
The whole algorithm of our proposed model is detailed in Algorithm \ref{alg:sml}.
Figure \ref{fig:comparison} visually illustrates the difference between transfer learning, model-agnostic meta-learning (MAML) and our proposed structured meta-learning (SML).
All methods in Figure \ref{fig:comparison} use tasks $\{\mathcal{T}_1, ..., \mathcal{T}_6 \}$ for training, $\mathcal{T}_{val}$ for validation and $\mathcal{T}_{test}$ for target task adaptation.
Transfer learning in Figure \ref{fig:comparison}(a) solves the training tasks $\{\mathcal{T}_1, ..., \mathcal{T}_6 \}$ in the multi-task learning approach without knowing the adaptation task $\mathcal{T}_{test}$. 
It aims at solving tasks in training set and select models based on the validation task $\mathcal{T}_{val}$.
MAML in Figure \ref{fig:comparison}(b) tries to learn transferable representation by repeatedly simulating the learning process in low-resource task-specific learning. In effect, the learned model can adapt fast to any unseen task, such as $\mathcal{T}_{test}$.
SML in Figure \ref{fig:comparison}(c) additionally explores the structure across tasks so that similar tasks will be initialized from closer starting points than others. In the testing (adaptation) stage, a new sentence function such as Negative Declarative will benefit from this learned structure by initializing from a point that is close to other fine-grained sentence functions in the same category of Declarative.

\section{Experimental Settings}
\paragraph{Dataset.}
We conduct experiments on STC-SeFun dataset \cite{Bi2019FineGrainedSF} which is a large-scale Chinese short text conversation dataset with manually labeled sentence functions.
Utterances in STC-SeFun have two-level sentence function labels. The four major sentence function types include: \textit{Declarative}, \textit{Interrogative}, \textit{Imperative} and \textit{Exclamatory}.
Each major sentence function is further divided into fine-grained sentence function labels like \textit{Wh-style Interrogative} which in total is 20 categories.
Considering all query and response sentence functions, we could have $20 \times 20 = 400$ meta tasks.
However, some tasks are extremely low-resource with less than 100 samples.
Incorporating these tasks as our adaptation tasks leads to a high variance of test performance.
To establish concrete evaluation, we only consider tasks with more than 700 samples, in which 100 samples are used for validation in the adaptation stage and 500 samples are used for the final testing.
Under this constraint, we receive 18 query-response fine-grained sentence function pairs as 18 tasks.
We select 9 high-resource tasks for meta-training, 4 tasks for meta-validation and 5 tasks for testing (adaptation).
The dataset statistics is shown in Table \ref{tab:dataset}.
Although 18 tasks are far smaller than thousands of tasks for few-shot image classification, it is still comparable to previous works such as low-resource machine translation \cite{Gu2018MetaLearningFL}.
Moreover, the experimental results may generalize to more tasks if there is more data.

\begin{table}[t!]
\fontsize{7.5}{9}\selectfont
\centering
\begin{tabular}{l|l|l|c|c|c}
\hline\hline
                                          & Query SF    & Response SF           & \multicolumn{3}{c}{\# Samples} \\
\hline
\multirow{9}{*}{\makecell{Meta \\ Train}}                    & Positive DE & Positive DE           & \multicolumn{3}{c}{27058}      \\
                                          & Wh-style IN & Positive DE           & \multicolumn{3}{c}{12854}      \\
                                          & Positive DE & Negative DE           & \multicolumn{3}{c}{5831}       \\
                                          & Negative DE & Positive DE           & \multicolumn{3}{c}{4006}       \\
                                          & Positive DE & Wh-style IN           & \multicolumn{3}{c}{3935}       \\
                                          & A-not-A IN  & Positive DE           & \multicolumn{3}{c}{3508}       \\
                                          & Wh-style IN & Negative DE           & \multicolumn{3}{c}{3367}       \\
                                          & Yes-no IN   & Positive DE           & \multicolumn{3}{c}{3267}       \\
                                          & Negative DE & Negative DE           & \multicolumn{3}{c}{2466}       \\
\hline
\multirow{4}{*}{\makecell{Meta \\ Val}}                      & Wh-style IN & DE w/ IN words      & 271 & 100 & 500        \\
                                          & Negative DE & Wh-style IN           & 161 & 100 & 500        \\
                                          & Positive DE & EX w/ interjections & 134 & 100 & 500        \\
                                          & Positive DE & DE w/ IN words      & 120 & 100 & 500        \\
\hline
\multirow{5}{*}{\makecell{Meta \\ Test}}                     & Positive DE & Yes-no IN             & 1314 & 100 & 500       \\
                                          & Yes-no IN   & Negative DE           & 893 & 100 & 500       \\
                                          & Positive DE & EX w/o tone words & 846 & 100 & 500       \\
                                          & A-not-A IN  & Negative DE           & 684 & 100 & 500       \\
                                          & Wh-style IN & Wh-style IN           & 488 & 100 & 500       \\
\hline\hline
\end{tabular}
\caption{Dataset statistics for our experiments. SF: Sentence Function; DE: Declarative; IN: Interrogative; EX: Exclamatory. For meta validation and meta test tasks, samples are further split into train, validation (100 samples) for task-specific adaptation. The rest 500 samples are used to test the performance of adapted models.}
\label{tab:dataset}
\end{table}

\paragraph{Baselines and Ablations.}
We train the model C-Seq2Seq described in Section \ref{sec:seq2seq} under the following learning settings:
\begin{itemize}[wide=0\parindent,noitemsep]
    \item Multi-Task Learning (MTL): We train C-Seq2Seq under the multi-task learning approach with training and validation data. Then we directly apply the trained model on each target task without fine-tuning.
    \item Multi-Task Learning + Fine-tuning (MTL+FT): We train C-Seq2Seq under the same training paradigm as multi-task learning. Then we fine-tune the model on each target task. This setting corresponds to a transfer learning scenario.
    \item Model-Agnostic Meta-Learning (MAML): We meta-train the model under the methodology model-agnostic meta-learning in other low-resource NLG tasks \cite{Gu2018MetaLearningFL,Qian2019DomainAD}. Then we fine-tune the meta-trained model on each target task. This is the ablation without structure modeling (Sec. \ref{sec:structure-maml}).
    \item Structured Meta-Learning (SML): We meta-train the model using structured meta-learning described in Algorithm \ref{alg:sml}. Then we fine-tune the meta-trained model on each target task.
\end{itemize}


\begin{table*}[ht!]
\small
\centering
\resizebox{1.0\textwidth}{!}{
\begin{tabular}{ll|l|cccc|cccccc}
\hline\hline
         \multirow{2}{*}{Query SF}            &      \multirow{2}{*}{Response SF}     & \multirow{2}{*}{Model}  & \multicolumn{4}{c|}{Human Evaluation}  & \multicolumn{6}{c}{Automatic Evaluation}\\
                    &                             &   & \texttt{Flue}  & \texttt{Rele}  & \texttt{Info}  & \texttt{Accu}  & \texttt{FT Step}   & \texttt{PPL}    & \texttt{B1}    & \texttt{B2}   & \texttt{Dist1} & \texttt{Dist2} \\
\hline
\multirow{4}{*}{Positive DE} & \multirow{4}{*}{Yes-no IN}             & MTL    & 59.40 & 50.40 & 36.53 & 0.33  & n/a    & 177.44 & 4.59  & 1.49 & 0.11  & 0.16  \\
                            &                                       & MTL+FT & 62.47 & 53.93 & \uline{39.87} & 34.00 & 304.00 & {91.65}  & 10.47 & 3.38 & 0.17  & 0.36  \\
                            &                                       & MAML   & \uline{63.40} & \uline{55.87} & \uline{39.87} & \uline{57.67} & 111.33 & 87.55  & {12.20} & {4.47} & {0.18}  & {0.39}  \\
                            &                                       & SML & \textbf{64.27} & \textbf{56.00} & \textbf{40.13} & \textbf{69.00} & \textbf{104.00} & {87.28}  & {12.88} & {4.72} & {0.21}  & {0.48}  \\
\hline
\multirow{4}{*}{Yes-no IN}   & \multirow{4}{*}{Negative DE}           & MTL    & 60.47 & 57.07 & 49.87 & 6.00  & n/a    & 73.04  & 5.46  & 1.46 & 0.10  & 0.14  \\
                            &                                       & MTL+FT & 61.07 & 56.80 & \uline{54.00} & 73.00 & 195.00 & {56.85}  & 9.98  & 3.37 & {0.22}  & {0.40}  \\
                            &                                       & MAML   & \uline{62.00} & \textbf{59.53} & 53.67 & \textbf{91.00} & \uline{111.00} & 58.79  & {11.12} & {3.81} & {0.16}  & {0.26}  \\
                        &                                           & SML & \textbf{64.93} & \uline{57.80} & \textbf{55.93} & \textbf{91.00} & \textbf{102.33} & {58.65}  & {11.91} & {3.76} & 0.13  & 0.20  \\
\hline
\multirow{4}{*}{Positive DE} & \multirow{4}{*}{\tabincell{c}{EX without \\ tone words}} & MTL    & 57.13 & 53.53 & 35.40 & 1.00  & n/a    & 100.21 & 6.06  & 2.69 & 0.14  & 0.23  \\
            &                       & MTL+FT & 56.40 & 53.67 & 36.67 & 39.00 & 268.00 & 80.32  & 12.20 & 4.26 & 0.12  & 0.22  \\
            &                       & MAML   & \uline{65.33} & \uline{56.20} & \uline{39.27} & \textbf{71.00} & \textbf{89.33}  & {68.08}  & {14.09} & {4.61} & {0.19}  & {0.40}  \\
            &                       & SML & \textbf{65.80} & \textbf{57.13} & \textbf{40.93} & \uline{68.00} & \uline{91.33}  & {71.71}  & {14.94} & {4.94} & {0.17}  & {0.33}  \\
\hline
\multirow{4}{*}{A-not-A IN}  & \multirow{4}{*}{Negative DE}           & MTL    & 60.33 & 54.93 & 49.73 & 4.33  & n/a    & 73.99  & 5.89  & 1.77 & 0.10  & 0.17  \\
        &                                                           & MTL+FT & 62.13 & 55.13 & \uline{51.47} & 53.67 & 158.00 & {59.53}  & 9.79  & 3.41 & {0.15}  & {0.27}  \\
            &                                                       & MAML   & \uline{62.60} & \uline{55.20} & 51.27 & \uline{89.33} & \uline{114.33} & 60.35  & {10.39} & {3.60} & {0.13}  & {0.20}  \\
        &                                                           & SML & \textbf{63.27} & \textbf{56.00} & \textbf{52.80} & \textbf{96.00} & \textbf{105.00} & {58.24}  & {11.28} & {3.96} & 0.12  & 0.19  \\
\hline
\multirow{4}{*}{Wh-style IN} & \multirow{4}{*}{Wh-style IN}           & MTL    & 62.47 & 51.67 & 38.33 & 1.00  & n/a    & 97.24  & 7.63  & 2.33 & 0.14  & 0.22  \\
                                            &                       & MTL+FT & 63.60 & 52.60 & 39.13 & 22.33 & 167.00 & 61.70  & {7.98}  & 2.58 & 0.17  & 0.30  \\
                                            &                       & MAML   & \uline{64.07} & \uline{53.13} & \uline{43.33} & \uline{85.00} & \uline{88.00}  & {44.02}  & 7.84  & {3.31} & {0.19}  & {0.46}  \\
                                            &                       & SML & \textbf{64.13} & \textbf{53.80} & \textbf{45.20} & \textbf{88.00} & \textbf{83.33}  & {43.06}  & {8.04}  & {3.96} & {0.19}  & {0.43}  \\
\hline\hline
\end{tabular}
}

\caption{Human evaluation results (in percentage \%) and automatic evaluation results in five testing tasks. The best/second-best results are bold/underlined except automatic metrics which are inconsistent with human perceptions~\cite{Liu2016HowNT}. Note that MTL does not fine-tune on target tasks, so FT Step is not applicable to this setting.}
\label{tab:main-result}
\end{table*}

\begin{table}[!t]
    \centering
    \resizebox{1.0\columnwidth}{!}{
    \begin{tabular}{l | c c c c }
    \Xhline{2\arrayrulewidth}
   Metrics &  \texttt{Flue} & \texttt{Rele} & \texttt{Info} & \texttt{Accu} \\
    \hline
    Fleiss's Kappa $\kappa$    & 0.61    & 0.72 & 0.67 & 0.90 \\
    \Xhline{2\arrayrulewidth}
    \end{tabular}
    }
    \caption{
    Fleiss's Kappa score for evaluating the inter annotator agreement.
    }
    \label{tab:result-k}
\end{table}

\paragraph{Model Settings.}
We take the most frequent 30k words as our vocabulary and use the pretrained embeddings \cite{Song2018DirectionalSE} for initialization. 
The sentence function embedding with dimension 20 is randomly initialized and learned through training.
We use two-layer LSTMs in both encoder and decoder, and the LSTMs hidden unit size is set to 400.
We use dropout \cite{Srivastava2014DropoutAS} with the probability $p=0.3$. 
All trainable parameters, except word embeddings, are randomly initialized with the uniform distribution in $(-0.1, 0.1)$. 
We adopt the teacher-forcing for the training.
In the testing, we select the model with the lowest perplexity and beam search with size 5 is employed for generation.
All hyper-parameters and models are selected on the validation dataset.

\paragraph{Learning Settings.}
We use SGD as the optimizer with a minibatch size of 64 and an initial learning rate of 1.0 for both meta-learning (line 9 and line 11 in Algorithm \ref{alg:sml}) and multi-task learning.
For meta-learning, we sample 3 tasks for line 3 in Algorithm \ref{alg:sml} and take a single gradient step for line 9 and line 11 in Algorithm \ref{alg:sml}. We meta-train the model for 8 epochs and start having the learning rate after the 3 epoch.
All models are fine-tuned with a SGD optimizer with a minibatch size of 64 and learning rate of 0.1.
We set the gradient norm upper bound to 3 and 1 during the training and fine-tuning respectively.
To avoid any random results, we report the average of five runs for all results.

\paragraph{Evaluation Metrics.}
Since automatic metrics for open-domain conversations may not be consistent with human perceptions \cite{Liu2016HowNT}, we hire 5 full-time human judges from a third-party data annotation company. We provide them a detailed annotation guideline (in Chinese) with good and bad response samples for each metric. They are first asked to annotate 100 responses for trial, and we select the top 3 judges according to the annotation quality. Finally, the 3 selected judges independently evaluate 2,000 responses generated from our model and three baselines for all five adaptation tasks. For each query, four responses generated from the proposed model, and three baselines are randomly shuffled to reduce the priming effect. 

The annotators evaluate responses on four metrics:
(1) ``Fluency'' (\texttt{Flue}) measures the grammatical correctness of responses;
(2) ``Relevance'' (\texttt{Rele}) measures whether the response is a relevant reply to the query;
(3) ``Informativeness'' (\texttt{Info}) evaluates whether the response provides any meaningful information with regard to the query;
(4) ``Accuracy'' (\texttt{Accu}) evaluates whether the response is coherent with the given response sentence function.
``Fluency'', ``Relevance'' and ``Informativeness'' are graded independently in a 1-5 scale where 5 is the best.
``Accuracy'' takes a binary value (1 or 0). 
We further normalize the average scores over all rated samples into $[0, 1]$.
Besides, we compare the adaption time of all models by calculating the ``Fine-Tuning Step'' (\texttt{FT Step}) till convergence on each test task.
For completeness, we also show the following automatic evaluation metrics: ``Perplexity'' (\texttt{PPL}); ``BLEU-1/2'' (\texttt{B1/B2})~\cite{Papineni2002BleuAM}; ``Distinct-1/2'' (\texttt{Dist1/Dist2})~\cite{li-etal-2016-diversity}.

\section{Results}
\paragraph{Performance of Human Evaluation.}
Human evaluation on ``Fluency'' (\texttt{Flue}), ``Relevance'' (\texttt{Rele}), ``Informativeness'' (\texttt{Info}) and ``Accuracy'' (\texttt{Accu}) are shown in Table \ref{tab:main-result}. The inter annotator agreement $\kappa$ scores are shown in Table \ref{tab:result-k}. We can make the following observations:
\begin{itemize}[noitemsep,leftmargin=*]
    \item MTL receives the worst performance on all human evaluation metrics. 
    Recall that MTL is trained on a mixture of all training tasks.
    It can only learn some generic response patterns like ``So do I''.
    That is why MTL has the worst performance on ``Informativeness'' metric.
    Moreover, since MTL never sees responses in target sentence functions in training, the generated responses are not coherent with the given sentence function at all.
    \item MTL+FT achieves better performance than MTL because it further fine-tunes on each target dataset. However, the performance on the accuracy of target sentence function is still unsatisfactory. This reveals that fine-tuning may not solve the adaptation problem on low-resource tasks.
    \item SML and MAML achieve the best/second-best human evaluation results across most of the metrics. This indicates that by simulating the low-resource testing scenarios in meta-training, the learned model adapts well on the low-resource testing tasks.
    Moreover, there is a huge improvement on the accuracy of given sentence functions (Accu), which reveals that MAML/SML can find model parameters that are sensitive to changes in the new task, such that small changes in the parameters produce large improvement on the accuracy of sentence functions.
    \item SML outperforms MAML in most of the cases. This tells us exploring the structure of sentence functions can balance knowledge generation and knowledge customization. The task-specific initialized model can leverage the knowledge of similar tasks and thus adapts the target tasks better.
\end{itemize}

\paragraph{Performance of Automatic Evaluation.}
We also show the results of automatic evaluations in Table \ref{tab:main-result}.
Compared to transfer learning based model MTL+FT, our meta-learning based models adapt faster (lower fine-tune step) and better (lower perplexity).
Although BLEU is not reliable enough to evaluate response generation, MAML and SML still achieve slightly better results than baselines.
Presumably, they can capture frequent word pattern in low-resource tasks.
Finally, MTL+FT, MAML and SML achieve comparable performance with regard to the unigram/bigram diversity (Dist1/2) of generated responses.

\paragraph{Effect of Structure Modeling.}
\begin{figure}[t!]
\centering
\includegraphics[width=1.0\columnwidth]{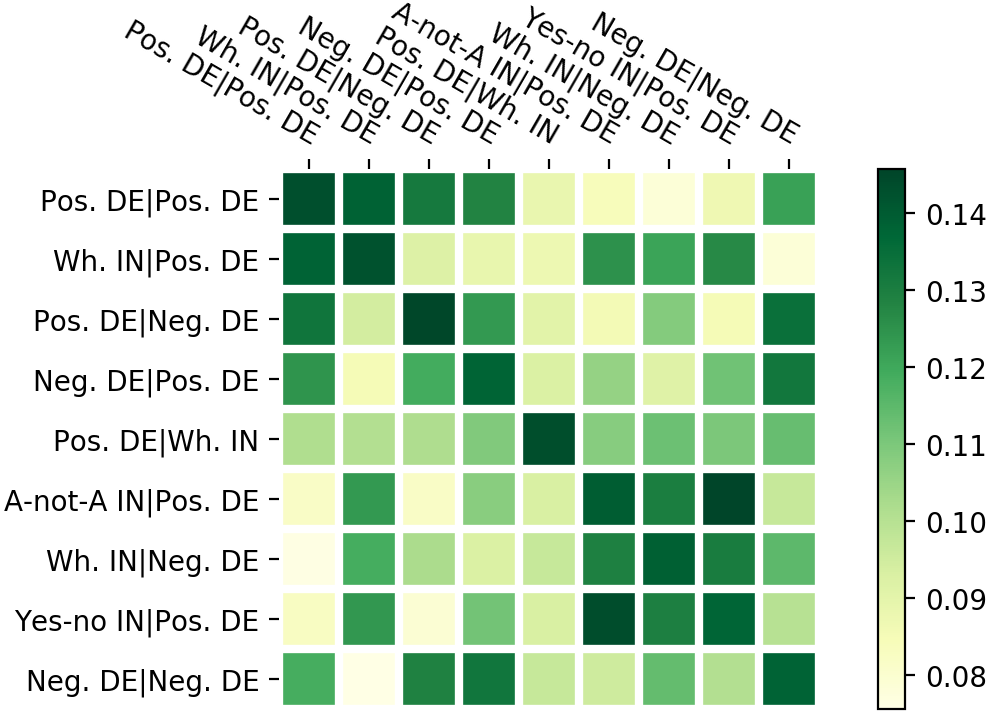}
\caption{Heatmap of the self-attention weight matrix. Each row shows the attention distribution $\mathbf{a}_k$ in Eqn.~\ref{eqn:selfattn1} for a given query-response sentence function pair (denoted in ``Query $|$ Response'' format).}
\label{fig:structure}
\end{figure}

To get more insight into how our proposed SML balances the knowledge generalization and customization, in Figure \ref{fig:structure}, we visualize the heatmap of self-attention weight $\mathbf{a}_k$ in Eqn.~\ref{eqn:selfattn1} for all 9 training sentence function representations.
Each row in Figure \ref{fig:structure} demonstrates the similarity between sentence function $\mathbf{s}_k$ and all sentence functions $[\mathbf{s}_1; ...; \mathbf{s}_K]$.
Take the first row in Figure \ref{fig:structure} for example, it tells us how all nine query-response sentence function pairs contribute to the representation of current query-response sentence function pair ``\textit{Positive Declarative (DE) $|$ Positive Declarative (DE)}''.
We see that sentence functions containing \textit{Interrogative(IN)} have nearly zero contribution while sentence functions containing \textit{Declarative(DE)} in both query and response have a certain amount of contribution.
Therefore, similar sentence functions trigger similar initializations and dissimilar sentence functions trigger different ones.

\paragraph{Case Study.}
\begin{figure}[t!]
\centering
\includegraphics[width=1.0\columnwidth]{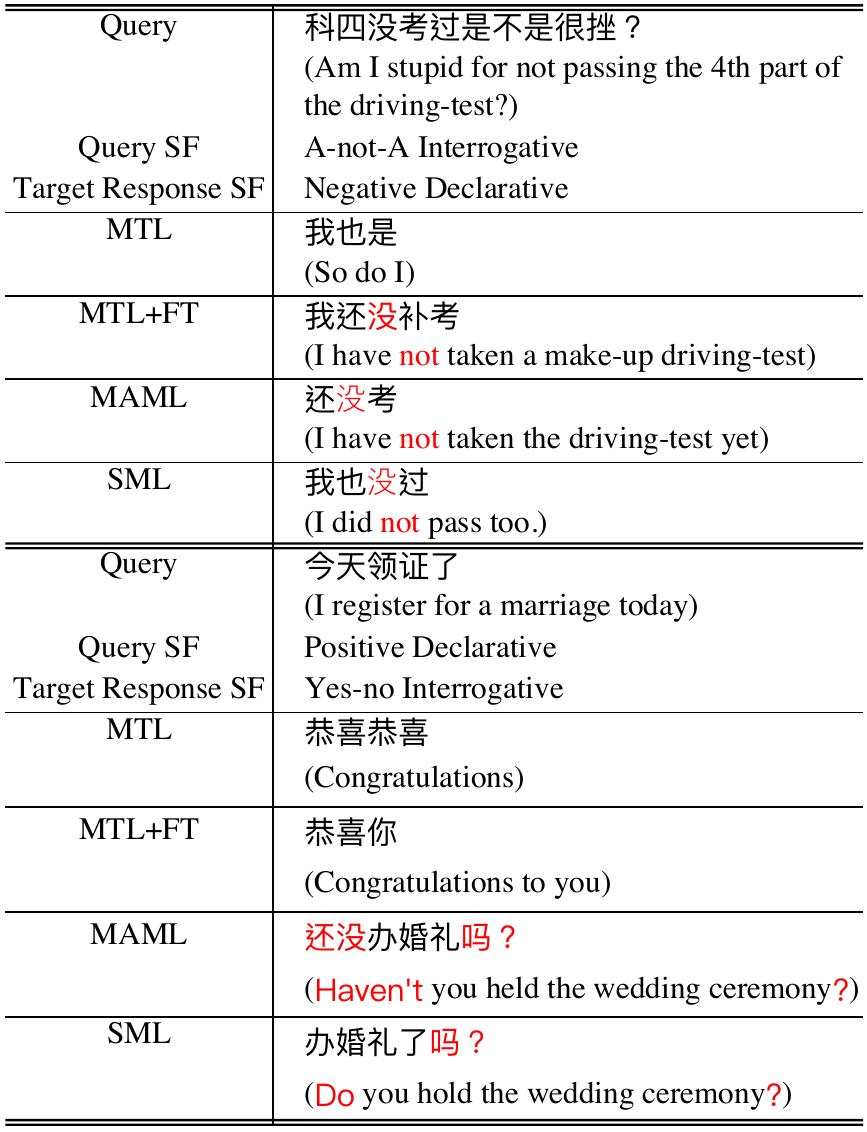}
\caption{Responses of all models. Words in red are related to the target sentence function. SF: Sentence Function.}
\label{fig:case}
\end{figure}
We present two examples in Figure \ref{fig:case}, each of which shows a test query with the target sentence function and responses generated by all models.
We see that responses generated by MTL are generic and can be used to reply to a large number of queries.
With fine-tuning on responses of the target sentence function, MTL+FT can capture the correct response pattern in some cases.
However, it is inferior to our proposed models MAML and SML, which can not only generate words related to the target sentence function but also keep the coherence and informativeness of responses.

\section{Conclusion}
In this paper, we propose a structured meta-learning algorithm for open domain dialogue generation on infrequent sentence functions.
To tackle the low-resource issue, our proposed model, based on the recently proposed model-agnostic meta-learning, can find both transferable internal representations and sensible parameters which can produce large improvement under a few adaptation steps.
Moreover, we further explore the structure across fine-grained sentence functions and such that the model can balance knowledge generalization and knowledge customization.
Extensive experiments show that our structured meta-learning (SML) algorithm outperforms existing approaches under the low-resource setting.

\section*{Acknowledgments}
The work described in this paper was partially supported by the National Key Research and Development Program of China (No. 2018AAA0100204) and the Research Grants Council of the Hong Kong Special Administrative Region, China (CUHK 14210717, RGC General Research Fund).

\bibliographystyle{acl_natbib}
\bibliography{emnlp2020}

\end{document}